\documentclass[11pt,a4paper]{article}
\pdfoutput=1
\usepackage[hyperref]{acl2020} 
\usepackage{times}
\usepackage{latexsym}

\usepackage{microtype}
\usepackage{paralist}
\aclfinalcopy 
\def\aclpaperid{3049} 
\title{Influence Paths for Characterizing Subject-Verb Number Agreement in LSTM Language Models}

\author{Kaiji Lu, Piotr Mardziel, Klas Leino, Matt Fedrikson, Anupam Datta \\
  Carnegie Mellon University}

\date{}

\usepackage{amsmath,amsfonts,bm}

\def\eqref#1{equation~\ref{#1}}

\def\1{\bm{1}}

\DeclareMathAlphabet{\mathsfit}{\encodingdefault}{\sfdefault}{m}{sl}
\SetMathAlphabet{\mathsfit}{bold}{\encodingdefault}{\sfdefault}{bx}{n}

\usepackage{url}
\usepackage{booktabs}       
\usepackage{amsfonts}       
\usepackage{nicefrac}       
\usepackage{microtype}      

\usepackage{subcaption}     

\usepackage{marvosym}
\usepackage{amssymb}
\usepackage{amsmath}
\usepackage{amsthm}
\usepackage{graphicx}
\usepackage{placeins}           
\usepackage{marginnote}
\usepackage{xspace}
\usepackage{tikz}
\usetikzlibrary{arrows,shapes,backgrounds,fit,positioning}
\graphicspath{{images/}}

\usepackage{multirow}

\usepackage{printlen}
\usepackage{relsize}

\usepackage{array}
\newcolumntype{x}[1]{>{\centering\arraybackslash}m{#1}}

\usepackage{footnote}     
\makesavenoteenv{tabular}
\makesavenoteenv{table}

\newtheorem{definition}{Definition}

\newcommand{\task}[1]{\text{#1}}

\newcommand{\compl}[1]{\overline{#1}} 

\usepackage{dsfont}

\usepackage{wasysym}

\newtheorem{observation}{Observation}

\newcommand{\stacklabel}[1]{\stackrel{\smash{\scriptscriptstyle \mathrm{#1}}}}
\newcommand{\defeq}{\stacklabel{def}=}

\newcommand{\seq}[1]{\mathbf #1}

\newcommand{\paren}[1]{\left( #1 \right)}

\newcommand{\vparen}[1]{\left< #1 \right>}
\newcommand{\set}[1]{\left\{ #1 \right\}}

\newcommand{\ra}{\rightarrow}

\newcommand{\rastar}[0]{\ra^*}

\def\final{}
\ifdefined\final
\else
\def\enablecomments
\fi

\ifdefined\draft
\def\enablecomments{}
\fi

\usepackage{soul}

\definecolor{LightGreen}{rgb}{0.80,1.00,0.80}
\definecolor{LightBlue}{rgb}{0.80,0.80,1.00}
\definecolor{LightRed}{rgb}{1.00,0.80,0.80}
\definecolor{LightPurple}{rgb}{0.94,0.85,1.00}
\definecolor{LightGray}{rgb}{0.90,0.90,0.90}

\soulregister{\method}{7}
\soulregister{\xspace}{7}
\soulregister{\emph}{7}

\ifdefined\enablecomments
  \DeclareRobustCommand{\commentformat}[3]{\sethlcolor{#2}\textsf{\hl{#1: #3}}}
  \newcommand{\sm}    [1]{{\scriptsize\sethlcolor{LightGray}\hl{\textsf{#1}}}}
\else
  \newcommand{\commentformat}[3]{}
  \newcommand{\sm}    [1]{}
\fi

\newcommand{\pxm}   [1]{\commentformat{PM}{LightGreen}{#1}}

\newcommand{\klas}  [1]{\commentformat{KL}{LightPurple}{#1}}

\newcommand{\todo}[1]{\commentformat{TODO}{LightRed}{#1}}

\usepackage{tikzit}
\usepackage{siunitx}

\tikzstyle{rec}=[] 
\tikzstyle{medium box}=[fill=white, draw=black, shape=circle, minimum height=0.40cm, inner sep=0mm]
\tikzstyle{Node}=[fill=black, draw=black, shape=circle, minimum height=0.01cm, inner sep=0.8mm]
\tikzstyle{Rectangle}=[fill=white, draw=black, shape=rectangle, tikzit shape=rectangle]
\tikzstyle{none}=[inner sep=0mm]

\tikzstyle{Edge}=[->, draw=black]
\tikzstyle{red edge}=[draw={rgb,255: red,191; green,0; blue,64}, ->, line width=1.6pt]
\tikzstyle{blue edge}=[draw={rgb,255: red,0; green,0; blue,255}, ->, line width=1.6pt]

\begin{document}

\maketitle


\begin{abstract}
  LSTM-based recurrent neural networks are the state-of-the-art for many
  natural language processing (NLP) tasks.
  Despite their performance, it is unclear whether, or how, LSTMs learn structural features of natural languages such as subject-verb number agreement in English.
  Lacking this understanding, the generality of LSTMs on this task and their suitability for related tasks remains uncertain. Further, errors cannot be properly attributed to a lack of structural capability, training data omissions, or other exceptional faults.
  We introduce \emph{influence paths}, a causal account
  of structural properties as carried by paths across gates and neurons of a recurrent neural
  network.
  The approach refines the notion of influence (the subject's grammatical number has influence on
  the grammatical number of the subsequent verb) into a set of gate-level or neuron-level paths.
  The set localizes and segments the concept (e.g., subject-verb agreement), its constituent
  elements (e.g., the subject), and related or interfering elements (e.g., attractors).
  We exemplify the methodology on a widely-studied multi-layer LSTM language model, demonstrating its accounting for subject-verb number agreement.
  The results offer both a finer and a more complete view of an LSTM's handling of this structural aspect of the English language than prior results based on diagnostic classifiers and ablation.

\end{abstract}

\section{Introduction}\label{sec:introduction}

\uselengthunit{in}

\begin{figure}[t]
  \begin{tikzpicture}[{every node/.style}={font=\small}]
	\begin{pgfonlayer}{nodelayer}
		\node [style=rec] (0) at (-14.5, -2) {In:};
		\node [style=rec] (out_label) at (-14.5, -3) {Out:};
		\node [style=rec] (1) at (-11, -2) {boys};
		\node [style=rec] (2) at (-8, -2) {behind};
		\node [style=rec] (3) at (-6, -2) {the};
		\node [style=rec] (4) at (-4, -2) {tree};
		\node [style=rec] (5) at (-2, -2) {(run)};
		\node [style=rec] (6) at (-13, -2) {The};

		\node [style=rec] (out_6) at (-13, -3) {$s_0$};
		\node [style=rec] (out_1) at (-11, -3) {$s_1$};
		\node [style=rec] (out_2) at (-8, -3) {$s_2$};
		\node [style=rec] (out_3) at (-6, -3) {$s_3$};
		\node [style=rec] (out_4) at (-4, -3) {$s_4$};
		
		\node [style=Rectangle] (7) at (-11, 0) {};
		\node [style=Rectangle] (8) at (-11, 2) {Cell $c_1^0$};
		\node [style=Rectangle] (9) at (-11, 6) {Candidate Cell $\tilde{c}_1^1$};
		\node [style=Rectangle] (10) at (-11, 8) {Cell $c_1^1$};
		\node [style=Rectangle] (11) at (-11, 0) {Candidate Cell $\tilde{c}_1^0$};
		\node [style=Rectangle] (12) at (-11, 4) {Hidden $h_1^0$};
		\node [style=Rectangle] (13) at (-8, 8) {$c_2^1$};
		\node [style=Rectangle] (14) at (-6, 8) {$c_3^1$};
		\node [style=Rectangle] (15) at (-4, 8) {$c_4^1$};
		\node [style=Rectangle] (17) at (-2, 4) {$h_4^1$};
		\node [style=Rectangle] (18) at (-4, 0) {};
		\node [style=Rectangle] (19) at (-4, 2) {$c_4^0$};
		\node [style=Rectangle] (20) at (-4, 6) {$\tilde{c}_4^1$};
		\node [style=Rectangle] (21) at (-4, 0) {$\tilde{c}_4^0$};
		\node [style=Rectangle] (22) at (-4, 4) {$h_4^0$};
		\node [style=Rectangle, align=center, text width=1in] (28) at (-3, -5) {agreement\\$s_4$(run) - $s_4$(runs)};
		\node [style=Rectangle, align=center, text width=1.2in] (29) at (-11, -5) {grammatical number\\ $ \text{boys} - \frac{\text{boy} + \text{boys}}{2}$};
	\end{pgfonlayer}
	\begin{pgfonlayer}{edgelayer}
		\draw [style=red edge] (1) to (11);
		\draw [style=red edge] (11) to (8);
		\draw [style=red edge] (8) to (12);
		\draw [style=red edge] (12) to (9);
		\draw [style=red edge] (9) to (10);
		\draw [style=red edge] (10) to (13);
		\draw [style=red edge] (13) to (14);
		\draw [style=red edge] (14) to (15);
		
		\draw [style=red edge,  bend left] (17) edge [transform canvas={xshift=-0.5mm}] (28);
		\draw [style=blue edge, bend left] (17) edge [transform canvas={xshift=+0.5mm}] (28);
		
		\draw [style=red edge,  bend left] (15) edge [transform canvas={xshift=-0.75mm}] (17);
		\draw [style=blue edge, bend left] (15) edge [transform canvas={xshift=+0.25mm, yshift=0.5mm}] (17);
		\draw [style=blue edge] (4) to (21);
		\draw [style=blue edge] (21) to (19);
		\draw [style=blue edge] (19) to (22);
		\draw [style=blue edge] (22) to (20);
		\draw [style=blue edge] (20) to (15);
		
		\draw [style=red edge, bend left] (29) edge (1);
		\coordinate (pre4) at (-5,-3);
		\draw [style=blue edge, bend left, dotted] (pre4) to (4);
		
		
	\end{pgfonlayer}
\end{tikzpicture}
  \caption{\label{fig:overview} Subject-verb agreement task for a 2-layer LSTM language model, and primary paths across various LSTM gates implementing subject-verb number agreement. A language model assigns score $ s $ to each word. Agreement is the score of the correctly numbered verb minus that of the incorrectly numbered verb.}\vspace{-.5em}
\end{figure}
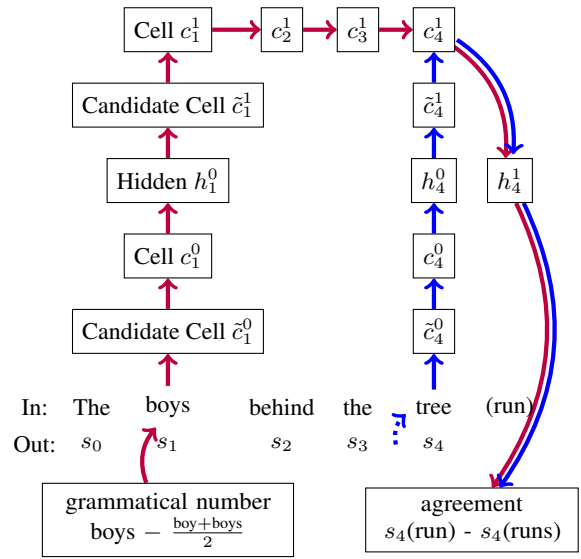



Traditional rule-based NLP techniques can capture syntactic structures, while statistical NLP
techniques, such as n-gram models, can heuristically integrate semantics of a natural language.
Modern RNN-based models such as Long Short-Term Memory (LSTM) models are tasked with incorporating
both semantic features from the statistical associations in their training corpus, and structural
features generalized from the same.

Despite evidence that LSTMs can capture syntactic rules in artificial languages~\cite{gers2001context}, it is
unclear whether they are as capable in natural
languages~\citep{linzen2016assessing, lakretz2019emergence} in the context of rules such as subject-verb number agreement, especially when not supervised for the particular feature. 
The incongruence derives from this central question: \emph{does an LSTM language model's apparent performance in
  subject-verb number agreement derive from statistical heuristics (like n-gram models) or from
  generalized knowledge (like rule-based models)?}

Recent work has begun addressing this question~\cite{linzen2016assessing} in the context of
\emph{language models}: models tasked with modeling the likelihood of the next word following a sequence of
words as expected in a natural language (see Figure~\ref{fig:overview}, bottom).
\emph{Subject-verb number agreement} dictates that the verb associated with a given subject should match
its number (e.g., in Figure~\ref{fig:overview}, the verb ``run'' should match with the subject ``boys'').
\citet{Giulianelli2018} showed that the subject grammatical number is associated with various gates in an
LSTM, and \citet{lakretz2019emergence} showed that ablation (disabling activation) of an LSTM model at certain
locations can reduce its accuracy at scoring verbs of the correct grammatical number.

\emph{Influence} offers an alternate means of exploring
properties like number agreement. We say an input is \emph{influential} on an outcome when changing
just the input and nothing else induces a change on the outcome. In English grammar, the number of
a subject is influential on the number of its verb, in that changing the number of that subject
while keeping all other elements of a sentence fixed would necessitate a change in the number of
the verb. Algorithmic transparency literature offers formal definitions for empirically quantifying notions of influence for systems in general~\cite{datta2016algorithmic} and for deep neural networks specifically~\cite{leino2018influence,sundararajan2017axiomatic}.

The mere fact that subject number is influential on verb number as output by an LSTM model is
sufficient to conclude that it incorporates the agreement concept in some way but does not indicate whether it operates as a statistical heuristic or as a generalized rule. We address this question with \emph{influence paths}, which decompose influence into a set of paths
across the gates and neurons of an LSTM model. The approach has several elements:
\begin{enumerate}
\item{} Define an input parameter to vary the concept-specific quantity under study (e.g., the
  grammatical number of a particular noun, bottom-left node in Figure~\ref{fig:overview}) and a concept-specific output feature to measure the
  parameter's effect on (e.g, number agreement with the parameterized noun, bottom-right node in Figure~\ref{fig:overview}).
\item{} Apply a gradient-based influence method to quantify the influence of the concept parameter
  on the concept output feature; as per the chain rule, decompose the influence into
  model-path-specific quantities.
\item{} Inspect and characterize the distribution of influence across the model paths.
\end{enumerate}

\noindent The paths demonstrate where relevant state information necessitated by the concept is
kept, how it gets there, how it ends up being used to affect the model's output, and how and where related
concepts interfere.

Our approach is state-agnostic in that it does not require \emph{a priori} an
assumption about how or if the concept will be implemented by the LSTM.
This differs from works on diagnostic classifiers where a representation of the concept is assumed to exist in the network's latent space.
The approach is also time-aware in that paths travel through cells/gates/neurons at different
stages of an RNN evaluation.
This differs from previous ablation-based techniques, which localize the number by clearing neurons at some position in an RNN
for all time steps.

%


Our contributions are as follows:
\begin{itemize}

\item{} We introduce \emph{influence paths}, a causal account of the use of concepts of interest
  as carried by paths across gates and neurons of an RNN.

\item{} We demonstrate, using influence paths, that in a multi-layer LSTM language model,
the concept of subject-verb number agreement is concentrated primarily on a single
  path (the red path in Figure~\ref{fig:overview}), despite a variety of surrounding and intervening contexts.

\item{} 
  We show that attractors (intervening nouns of
  opposite number to the subject) do not diminish the contribution of the primary subject-verb path,
  but rather contribute their own influence of the opposite direction along the equivalent primary
  attractor-verb path (the blue path in the figure). This can lead to incorrect number prediction if an attractor's contribution
  overcomes the subject's.

\item{} We corroborate and elaborate on existing results localizing subject number to the same two
  neurons which, in our results, lie on the primary path. We further extend and generalize prior
  compression/ablation results with a new path-focused compression test which verifies our
  localization conclusions.

\end{itemize}

\noindent Our results point to generalized knowledge as the answer to the central question. The
number agreement concept is heavily centralized to the primary path despite the varieties of
contexts. Further, the primary path's contribution is undiminished even amongst interfering contexts;
number errors are not attributable to lack of the general number concept but rather to sufficiently
influential contexts pushing the result in the opposite direction.


\section{Background}\label{sec:background}




\paragraph{LSTMs}




Long short-term memory networks (LSTMs)~\citep{hochreiter1997long} have proven to be effective for
modeling sequences, such as language models, and empirically, this architecture has been found to be optimal compared to other second-order RNNs~\citep{greff2017lstm}. LSTMs utilize several types of gates and internal states including \emph{forget gates} ($f$), \emph{input gates} ($i$), \emph{output gates} ($o)$, \emph{cell states} ($c$), \emph{candidate cell state} ($\tilde{c}$), and \emph{hidden states} ($h$).
Each gate is designed to carry out a certain function, or to fix a
certain drawback of the vanilla RNN architecture. 
E.g., the forget gate is supposed to determine
how much information from the previous cell state to retain or ``forget'', helping to fix the
vanishing gradient problem~\citep{hochreiter1998vanishing}.
  \pxm{Talk about sparse vs dense connections, name the various gate types and their
    function.}

  \pxm{Mention embedding and decoding and that we will work with embedded word vectors in this
    paper.}


 \paragraph{Number Agreement in Language Models}
The \emph{number agreement} (NA) task, as described by \citet{linzen2016assessing}, is an
evaluation of a language model's ability to properly match the verb's grammatical number with its subject. This evaluation is performed on sentences specifically designed for the exercise,
with zero or more words between the subject and the main verb, termed the \emph{context}. The task
for sentences with non-empty contexts will be referred to as \emph{long-term} number agreement.
\klas{We never actually specified what the task was! There should be a brief sentence doing this.}

``Human-level'' performance for this task can be achieved with a 2-layer LSTM language model~\citep{Gulordavaa}, indicating that the language model incorporates grammatical number despite being trained only for the more general word prediction task.
Attempts to explain or localize the number concept within the model include \citep{lakretz2019emergence}, where ablation of neurons is applied to locate specific neurons where such information is stored; and \cite{Giulianelli2018, hupkes2018visualisation}, where diagnostic classifiers are trained on gate activations to predict the number of the subject to see which gates or timesteps the number concept exhibits itself. These works also look at the special cases involving \emph{attractors}---intervening nouns with grammatical number opposite to that of the subject (deemed instead \textit{helpful nouns} if their number agrees with the subject)---such as the word ``tree'' in Figure~\ref{fig:overview}. Both frameworks provide explanations as to why attractors lower the performance of NA tasks. However, they tend to focus on the activation patterns of gates or neurons without justifying their casual relationships with the concept of grammatical number, and do not explicitly identify the exact temporal trajectory of how the number of the subject influences the number of the verb.

Other relevant studies that look inside RNN models to locate specific linguistic concepts include visualization techniques such as \citep{karpathy2015visualizing}, and explanations for supervised tasks involving LSTMs such as sentiment analysis \citep{murdoch2018beyond}.


\paragraph{Attribution Methods}

\emph{Attribution methods} quantitatively measure the contribution of each of a function's
individual inputs to its output. Gradient-based attribution methods compute the gradient of a model
with respect to its inputs to describe how important each input is towards the output predictions.
These methods have been applied to assist in explaining deep neural networks, predominantly in the
image domain~\citep{leino2018influence, sundararajan2017axiomatic,bach2015pixel, simonyan13saliency}. Some such
methods are also axiomatically justified to provide a causal link between inputs (or intermediate
neurons) and the output.


As a starting point in this work, we consider \emph{Integrated Gradients}
(IG)~\citep{sundararajan2017axiomatic}. Given a \emph{baseline},
$x_0$, the attribution for each input at point, $x$, is the path integral taken from the baseline to $x$ of the gradients of the
model's output with respect to its inputs.
The baseline establishes a neutral point from which to make a counterfactual comparison; the
attribution of a feature can be interpreted as the share of the model's output that is due to that
feature deviating from its baseline value. By integrating the gradients along the linear
interpolation from the baseline to $x$, IG ensures that the attribution given to each feature is
\emph{sensitive} to effects exhibited by the gradient at any point between the baseline and
instance $ x $.
%

\citet{leino2018influence} generalize IG to better focus attribution on concepts other than just
model outputs, by use of a \emph{quantity of interest} (QoI) and a \emph{distribution of interest}
(DoI). 
Their measure, \emph{Distributional Influence}, is given by Definition~\ref{def:distributional_inf}.
The QoI is a function of the model's output expressing a particular output behavior of the
model to calculate influence for; in IG, this is fixed as the model's
output. The DoI specifies a distribution over which the influence should
faithfully summarize the model's behavior; the influences are found by taking an expected value over DoI.

\begin{definition}[Distributional Influence]\label{def:distributional_inf}
With \emph{quantity of interest}, $q$, and \emph{distribution of interest}, $D$, the influence, $\chi$, of the inputs on the quantity of interest is:
\begin{equation*}
\chi(q, D) = \mathop{\mathlarger{\mathlarger{\mathbb{E}}}}_{\vec{x}\sim D}\left[\frac{\partial q}{\partial x}(\vec{x})\right]
\end{equation*}

\end{definition}


\noindent The directed path integral used by IG can be implemented by setting the DoI to a uniform distribution over the
line from the baseline to $\vec{x}$: $D=\text{Uniform}\paren{\overline{\vec{x}_0 \vec{x}}}$, for baseline, $\vec{x}_0$, and then multiplying $\chi$  by ${\vec{x} - \vec{x}_0}$.
Conceptually, by multiplying by $\vec{x} - \vec{x_0}$, we are measuring the \emph{attribution}, i.e., the contribution to the QoI, of $\vec{x} - \vec{x_0}$ by weighting its features by their \emph{influence}.
We use the framework of \citeauthor{leino2018influence} in this way to define our measure of attribution for NA tasks in Section~\ref{sec:methods}.

Distributional Influence can be approximated by sampling according to the DoI.
In particular, when using $D=\text{Uniform}\paren{\overline{\vec{x}_0 \vec{x}}}$ as noted above,  Definition~\ref{def:distributional_inf} can be computationally approximated with a sum of $n$ intervals as in IG:
\begin{equation*}
  \chi \approx \sum_{i=1}^n
  \frac{\partial q}{\partial x} \paren{\frac{i}{n} \vec{x} + \paren{1-\frac{i}{n}} \vec{x}_0}
\end{equation*}

\noindent Other related works include \citet{fiacco2019deep}, which employs the concept of neuron paths based on cofiring of neurons instead of influence, also on different NLP tasks from ours. 

\section{Methods}\label{sec:methods}
Our method for computing influence paths begins with modeling a relevant concept, such as grammatical number, in the influence framework of~\citeauthor{leino2018influence} (Definition~\ref{def:distributional_inf}) by defining a quantity of interest that corresponds to the grammatical number of the verb, and defining a component of the input embedding that isolates the subject's grammatical number 
(Section~\ref{sec:methods-concept}). We then decompose the influence measure along the relevant structures of LSTM (gates or neurons) as per
standard calculus identities to obtain a definition for \emph{influence paths} (Section~\ref{sec:methods-decomposition}). 

\subsection{Measuring Number Agreement}\label{sec:methods-concept}
For the NA task, we view the initial fragment containing the subject as the input, and the word
distribution at the position of its corresponding verb as the output.

Formally, each instance in this task is a sequence of $d$-dimensional word embedding vectors, $\seq{w} \defeq \vparen{\vec{w}_i}_i$, containing the subject and the corresponding
verb, potentially with intervening words in between. We assume the subject is at position $t$ and
the verb at position ${t+n}$. The output score of a word, $ w $, at position $ i $ will be written $
s_i(w) $. If $ w $ has a grammatical number, we write $ w^+ $ and $ w^- $ to designate $w$ with its original number and the equivalent word with the opposite number, respectively.

\paragraph{Quantity of Interest}
We instrument the output score with a QoI measuring the agreement of the output's
grammatical number to that of the subject:
\begin{definition}[Number Agreement Measure]\label{def:agreement}
  Given a sentence, $ \seq{w} $, with verb, $w$, whose correct form (w.r.t. grammatical number) is $w^+$, the \emph{quantity of interest}, $ q $, measures the correctness of the grammatical number of the verb:
  \begin{equation*}
    q\paren{\seq{w}} \defeq s_{t+n}\paren{w^+} - s_{t+n}\paren{w^-}
  \end{equation*}
\end{definition}

\noindent In plain English, $q$ captures the weight that the model assigns to the correct form of $w$ as opposed to the weight it places on the incorrect form.
Note that the number agreement concept could have reasonably been measured using a different quantity of interest. 
E.g., considering the scores of \emph{all} vocabulary words of the correct number and incorrect number in the positive and negative
terms, respectively, is an another alternative. However, based on our preliminary experiments, we found this
alternative does not result in meaningful changes to the reported results in the further sections.

\paragraph{Distribution of Interest}

We also define a component of the embedding of the subject that captures its grammatical number, and a distribution over the inputs that allows us to sensitively measure the influence of this concept on our chosen quantity of interest.
Let $ \vec{w}^0 $ be the word embedding
midway between its numbered variants, i.e., $ \frac{\vec{w}^+ + \vec{w}^-}{2} $. Though this vector will typically
not correspond to any English word, we interpret it as a number-neutral version of $ \vec{w} $. Various
works show that linear arithmetic on word embeddings of this sort preserves meaningful word
semantics as demonstrated in analogy parallelograms \cite{mikolov2013distributed}. Finally, given a sentence, $
\seq{w} $, let $ \seq{w}_t^0 $ be the sentence $ \seq{w} $, except with the word embedding $ \vec{w}_{t} $ replaced
with its neutral form $ \vec{w}_t^0 $.
We see that $\seq{w} - \seq{w}_t^0$ captures the part of the input corresponding to the grammatical number of the subject, $ \vec{w}_{t} $. 

\begin{definition}[Grammatical Number Distribution]\label{def:distribution}
  Given a singular (or plural) noun, $ w_t $, in a sentence, $ \seq{w} $, the distribution density of
  sentences, $ D_\seq{w} $, exercising the noun's \emph{singularity} (or \emph{plurality}) linearly interpolates
  between the neutral sentence, $ \seq{w}_t^0 $, and the given sentence, $ \seq{w} $:
  \begin{equation*}
    D_\seq{w} \defeq \text{Uniform}\paren{\overline{\seq{w}_t^0 \seq{w}}}
  \end{equation*}

\end{definition}

\noindent If $ \vec{w}_t $ is singular, our counterfactual sentences span $ \seq{w} $ with number-neutral $ \vec{w}^0_t $
all the way to its singular form $ \vec{w}_t = \vec{w}^+_t $. We thus call this distribution a
\emph{singularity} distribution. Were $ w_t $ plural instead, we would refer to the
distribution as a \emph{plurality} distribution. Using this distribution of sentences as our DoI
thus allows us to measure the influence of $\seq{w} - \seq{w}_t^0$ (the grammatical number of a noun at position $ t $) on
our quantity of interest \emph{sensitively} (in the sense that \citeauthor{sundararajan2017axiomatic} define their axiom of sensitivity for IG
~\cite{sundararajan2017axiomatic}). 

\paragraph{Subject-Verb Number Agreement}
Putting things together, we define our attribution measure.

\begin{definition}[Subject-Verb Number Agreement Attribution]\label{def:agreement-attribution}
  The measure of attribution, $\alpha$, of a noun's grammatical number on the subject-verb number agreement is
  defined in terms of the DoI, $ D_\seq{w} $, and QoI, $ q $, as
  in Definitions~\ref{def:distribution} and \ref{def:agreement}, respectively. 
%
%
  \begin{align*}\label{eq:number}
    \alpha\paren{\seq{w}} &= (\seq{w} - \seq{w}_t^0)~ \chi(q, D_\seq{w})
  \end{align*}

\end{definition}
\noindent Essentially, the attribution measure weights the features of the subject's grammatical number by their Distributional Influence, $\chi$.
Because $D_{\seq{w}}$ is a uniform distribution over the line segment between $\seq{w}$ and $\seq{w}_t^0$, as with IG, the attribution can be interpreted as each feature's net contribution to the change in the QoI, $q(\seq{w}) - q(\seq{w}_t^0)$, as $\sum_{i}{\chi(\seq{w})_i} = q(\seq{w}) - q(\seq{w}_t^0)$ (i.e., Definition~\ref{def:agreement-attribution} satisfies the axiom \citeauthor{sundararajan2017axiomatic} term \emph{completeness}~\cite{sundararajan2017axiomatic}).

In Figure \ref{fig:overview}, for instance, this definition measures the attribution
from the plurality of the subject (``boys''), towards the model's prediction of the
correctly numbered verb (``run'') versus the incorrectly numbered verb (``runs''). Later in
this paper we will also investigate the attribution of intervening nouns on this same quantity.
We expect the input attribution to be positive for all subjects and helpful nouns, and negative for
attractors, which can be verified by the $P^+$columns of Table~\ref{tab:primary} (the details of
this experiment are introduced in Section~\ref{sec:evaluation}).

\subsection{Influence Paths}\label{sec:methods-decomposition}

Input attribution as defined by IG~\citep{sundararajan2017axiomatic} provides a way of
explaining a model by highlighting the input dimensions with large attribution towards the output.
Distributional Influence~\citep{leino2018influence} with a carefully chosen QoI
and DoI (Definition~\ref{def:agreement-attribution}) further focuses the influence on a concept at hand, grammatical number agreement. Neither,
however, demonstrate how these measures are conveyed by the inner workings of a model. In this
section we define a decomposition of the influence into paths of a model, thereby assigning
attribution not just to inputs, but also to the internal structures of a given model.
\newcommand{\attrib}[0]{\mathcal{A}}
\newcommand{\attribst}[0]{\attrib_{s \ra t}}

We first define arbitrary deep learning models as computational graphs, as in Definition~\ref{def:model}.
We then use this graph abstraction to define a notion of influence for a path through the graph.
We posit that any natural path decomposition should satisfy the following conservation property: 
\emph{the sum of the influence of each path from the input to the output should equal the influence of the input on the QoI}.
We then observe that the chain rule from calculus offers one such natural decomposition, yielding  Definition~\ref{def:decomposition}.

\begin{definition}[Model]\label{def:model}
  A model is an acyclic graph with a set of nodes, edges, and activation functions associated with
  each node. The output of a node, $ n $, on input $ x $ is $ n(x) \defeq f_n\paren{n_1(x), \cdots,
    n_m(x)} $ where $ n_1, \cdots, n_m $ are $n$'s predecessors and $f_n $ is its activation
  function. If $ n $ does not have predecessors (it is an input), its activation is $ f_n(x) $.
  We assume that the domains and ranges of
  all activation functions are real vectors of arbitrary dimension.

  We will write $ n_1 \ra n_2 $ to denote an edge (i.e., $n_1$ is a direct predecessor of $n_2$), and $ n_1 \ra^* n_2 $ to
  denote the set of all paths from $ n_1 $ to $ n_2 $. The partial derivative of the activation of
  $n_2 $ with respect to the activation of $n_1 $ will be written $ \frac{\partial n_2}{\partial
    n_1} $.

\end{definition}
This view of a computation model is an extension of network decompositions from attribution methods using the natural concept of ``layers'' or ``slices''~\cite{dhamdhere2018important,leino2018influence,
 bach2015pixel}. This decomposition can be tailored to the level of granularity we wish to expose.
Moreover, in RNN models where no single and consistent ``natural layer'' can be found due to the
variable-length inputs, a more general graph view provides the necessary versatility.

\begin{definition}[Path Influence]
\label{def:decomposition}

  Expanding Definition~\ref{def:agreement-attribution} using the chain rule, the influence of input node, $ s $,
  on target node, $ t $, in a model, $ G $, is: \pxm{At what input?}
  \begin{align*}
    \chi_s & = \mathop{\mathlarger{\mathlarger{\mathbb{E}}}}_{x\sim D(x)}\left[\frac{\partial t}{\partial s}(x)\right]\\
          & = \mathop{\mathlarger{\mathlarger{\mathbb{E}}}}_{x\sim D(x)}\left[
      \sum_{p \in \paren{s \rastar t}} \prod_{\paren{n_1 \ra n_2} \in p} \frac{\partial n_2}{\partial n_1}(x)\right]\\
& = \sum_{p \in \paren{s \rastar t}} \underbrace{\mathop{\mathlarger{\mathlarger{\mathbb{E}}}}_{x\sim D(x)}\left[
      \prod_{\paren{n_1 \ra n_2} \in p} \frac{\partial n_2}{\partial n_1}(x)\right]}_{\chi_s^p}
  \end{align*}
\end{definition}

\noindent Note that the same LSTM can be modeled with different graphs to achieve a desired level of abstraction. 
We will use two particular levels of granularity: a coarse
\emph{gate-level} abstraction where nodes are LSTM gates, and a fine \emph{neuron-level} abstraction where nodes are
the vector elements of those gates. Though the choice of abstraction granularity has no effect
on the represented model semantics, it has implications on graph paths and the scale of their
individual contributions in a model.

\todo{axiom 1: Path invariance, if two path are
  functionally the same, they should have same influence;}

\todo{axiom 2: conservation of influence, the sum of all influence should equal to
  the total influence.}

\todo{uniqueness result for equation 5}

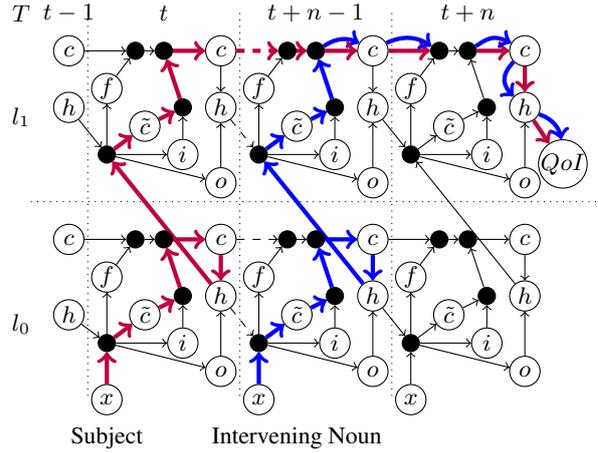
\begin{figure}

  \begin{tikzpicture}[{every node/.style}={font=\small}]
 \clip(-15.5,5) rectangle (-2,13);

	\begin{pgfonlayer}{nodelayer}
		\node [style=medium box] (1) at (-14, 7) {$c$};
		\node [style=medium box] (8) at (-14, 5) {$h$};
		\node [style=medium box] (10) at (-12, 5) {$\tilde{c}$};
		\node [style=medium box] (11) at (-11, 4.25) {$i$};
		\node [style=Node] (12) at (-12.25, 7) {};
		\node [style=Node] (13) at (-13, 4.25) {};
		\node [style=medium box] (14) at (-13, 2.75) {$x$};
		\node [style=Node] (15) at (-11, 5.5) {};
		\node [style=Node] (16) at (-11.5, 7) {};
		\node [style=medium box] (17) at (-10, 3.5) {$o$};
		\node [style=medium box] (18) at (-13, 6) {$f$};
		\node [style=medium box] (19) at (-10, 7) {$c$};
		\node [style=medium box] (21) at (-10, 5.5) {$h$};
		\node [style=none] (88) at (-13, 1.75) {Subject};
		\node [style=none] (88) at (-8, 1.75) {Intervening Noun};
		\node [style=medium box] (95) at (-8, 5) {$\tilde{c}$};
		\node [style=medium box] (96) at (-7, 4.25) {$i$};
		\node [style=Node] (97) at (-8.25, 7) {};
		\node [style=Node] (98) at (-9, 4.25) {};
		\node [style=medium box] (99) at (-9, 2.75) {$x$};
		\node [style=Node] (100) at (-7, 5.5) {};
		\node [style=Node] (101) at (-7.5, 7) {};
		\node [style=medium box] (102) at (-6, 3.5) {$o$};
		\node [style=medium box] (103) at (-9, 6) {$f$};
		\node [style=medium box] (104) at (-6, 7) {$c$};
		\node [style=medium box] (105) at (-6, 5.5) {$h$};
		\node [style=medium box] (106) at (-4, 5) {$\tilde{c}$};
		\node [style=medium box] (107) at (-3, 4.25) {$i$};
		\node [style=Node] (108) at (-4.25, 7) {};
		\node [style=Node] (109) at (-5, 4.25) {};
		\node [style=medium box] (110) at (-5, 2.75) {$x$};
		\node [style=Node] (111) at (-3, 5.5) {};
		\node [style=Node] (112) at (-3.5, 7) {};
		\node [style=medium box] (113) at (-2, 3.5) {$o$};
		\node [style=medium box] (114) at (-5, 6) {$f$};
		\node [style=medium box] (115) at (-2, 7) {$c$};
		\node [style=medium box] (116) at (-2, 5.5) {$h$};
		\node [style=medium box] (117) at (-14, 12) {$c$};
		\node [style=medium box] (118) at (-14, 10.5) {$h$};
		\node [style=medium box] (119) at (-12, 10) {$\tilde{c}$};
		\node [style=medium box] (120) at (-11, 9.25) {$i$};
		\node [style=Node] (121) at (-12.25, 12) {};
		\node [style=Node] (122) at (-13, 9.25) {};
		\node [style=Node] (124) at (-11, 10.5) {};
		\node [style=Node] (125) at (-11.5, 12) {};
		\node [style=medium box] (126) at (-10, 8.5) {$o$};
		\node [style=medium box] (127) at (-13, 11) {$f$};
		\node [style=medium box] (128) at (-10, 12) {$c$};
		\node [style=medium box] (129) at (-10, 10.5) {$h$};
		\node [style=medium box] (130) at (-8, 10) {$\tilde{c}$};
		\node [style=medium box] (131) at (-7, 9.25) {$i$};
		\node [style=Node] (132) at (-8.25, 12) {};
		\node [style=Node] (133) at (-9, 9.25) {};
		\node [style=Node] (135) at (-7, 10.5) {};
		\node [style=Node] (136) at (-7.5, 12) {};
		\node [style=medium box] (137) at (-6, 8.5) {$o$};
		\node [style=medium box] (138) at (-9, 11) {$f$};
		\node [style=medium box] (139) at (-6, 12) {$c$};
		\node [style=medium box] (140) at (-6, 10.5) {$h$};
		\node [style=medium box] (141) at (-4, 10) {$\tilde{c}$};
		\node [style=medium box] (142) at (-3, 9.25) {$i$};
		\node [style=Node] (143) at (-4.25, 12) {};
		\node [style=Node] (144) at (-5, 9.25) {};
		\node [style=Node] (146) at (-3, 10.5) {};
		\node [style=Node] (147) at (-3.5, 12) {};
		\node [style=medium box] (148) at (-2, 8.5) {$o$};
		\node [style=medium box] (149) at (-5, 11) {$f$};
		\node [style=medium box] (150) at (-2, 12) {$c$};
		\node [style=medium box] (151) at (-2, 10.5) {$h$};
		\node [style=medium box] (153) at (-1, 9) {$QoI$};
		\node [style=none] (154) at (-9.5, 13) {};
		\node [style=none] (155) at (-9.5, 2.75) {};
		\node [style=none] (156) at (-5.5, 13) {};
		\node [style=none] (157) at (-5.5, 2.75) {};
		\node [style=none] (158) at (-15, 8) {};
		\node [style=none] (159) at (0, 8) {};
		\node [style=none] (168) at (-15.25, 4.75) {$l_0$};
		\node [style=none] (169) at (-15.25, 10.25) {$l_1$};
		\node [style=none] (170) at (-15.25, 13) {$T$};
		\node [style=none] (171) at (-11.5, 13) {$t$};
		\node [style=none] (172) at (-7.5, 13) {$t+n-1$};
		\node [style=none] (173) at (-3.5, 13) {$t+n$};
		\node [style=none] (174) at (-13.5, 13) {};
		\node [style=none] (175) at (-13.5, 2.75) {};
		\node [style=none] (176) at (-14, 13) {$t-1$};
	\end{pgfonlayer}
	\begin{pgfonlayer}{edgelayer}
		\draw [style=Edge] (1) to (12);
		\draw [style=Edge] (8) to (13);
		\draw [style=Edge] (13) to (11);
		\draw [style=Edge] (11) to (15);
		\draw [style=Edge] (12) to (16);
		\draw [style=Edge] (13) to (18);
		\draw [style=Edge] (18) to (12);
		\draw [style=Edge] (13) to (17);
		\draw [style=Edge] (17) to (21);
		\draw [style=red edge] (14) to (13);
		\draw [style=red edge] (13) to (10);
		\draw [style=red edge] (10) to (15);
		\draw [style=red edge] (15) to (16);
		\draw [style=red edge] (16) to (19);
		\draw [style=red edge] (19) to (21);
		\draw [style=Edge] (98) to (96);
		\draw [style=Edge] (96) to (100);
		\draw [style=Edge] (97) to (101);
		\draw [style=Edge] (98) to (103);
		\draw [style=Edge] (103) to (97);
		\draw [style=Edge] (98) to (102);
		\draw [style=Edge] (102) to (105);
		\draw [style=Edge] (109) to (107);
		\draw [style=Edge] (107) to (111);
		\draw [style=Edge] (108) to (112);
		\draw [style=Edge] (109) to (114);
		\draw [style=Edge] (114) to (108);
		\draw [style=Edge] (109) to (113);
		\draw [style=Edge] (113) to (116);
		\draw [style=Edge] (117) to (121);
		\draw [style=Edge] (118) to (122);
		\draw [style=Edge] (122) to (120);
		\draw [style=Edge] (120) to (124);
		\draw [style=Edge] (121) to (125);
		\draw [style=Edge] (122) to (127);
		\draw [style=Edge] (127) to (121);
		\draw [style=Edge] (122) to (126);
		\draw [style=Edge] (126) to (129);
		\draw [style=red edge] (122) to (119);
		\draw [style=red edge] (119) to (124);
		\draw [style=red edge] (124) to (125);
		\draw [style=red edge] (125) to (128);
		\draw [style=Edge] (128) to (129);
		\draw [style=Edge] (133) to (131);
		\draw [style=Edge] (131) to (135);
		\draw [style=Edge] (133) to (138);
		\draw [style=Edge] (138) to (132);
		\draw [style=Edge] (133) to (137);
		\draw [style=Edge] (137) to (140);
		\draw [style=Edge] (144) to (142);
		\draw [style=Edge] (142) to (146);
		\draw [style=Edge] (143) to (147);
		\draw [style=Edge] (144) to (149);
		\draw [style=Edge] (149) to (143);
		\draw [style=Edge] (144) to (148);
		\draw [style=Edge] (148) to (151);
		\draw [style=red edge] (147) to (150);
		\draw [dashed][style=red edge] (128) to (132);
		\draw [style=red edge] (139) to (143);
		\draw [dotted](154.center) to (155.center);
		\draw [dotted](156.center) to (157.center);
		\draw [dotted](158.center) to (159.center);
		\draw [style=blue edge, bend left, looseness=1.25] (147) to (150);
		\draw [style=blue edge, bend left] (139) to (143);
		\draw [style=blue edge] (99) to (98);
		\draw [style=blue edge] (98) to (95);
		\draw [style=blue edge] (95) to (100);
		\draw [style=blue edge] (104) to (105);
		\draw [style=blue edge] (105) to (133);
		\draw [style=blue edge] (133) to (130);
		\draw [style=blue edge] (130) to (135);
		\draw [style=blue edge] (135) to (136);
		\draw [style=blue edge, bend left] (136) to (139);
		\draw [style=red edge] (21) to (122);
		\draw [dashed][style=Edge] (19) to (97);
		\draw [dashed][style=Edge] (21) to (98);
		\draw [style=Edge] (104) to (108);
		\draw [style=Edge] (105) to (109);
		\draw [style=Edge] (112) to (115);
		\draw [style=Edge] (109) to (106);
		\draw [style=Edge] (106) to (111);
		\draw [style=Edge] (111) to (112);
		\draw [style=Edge] (110) to (109);
		\draw [style=Edge] (116) to (144);
		\draw [style=Edge] (146) to (147);
		\draw [style=blue edge] (100) to (101);
		\draw [style=blue edge] (101) to (104);
		\draw [style=red edge] (136) to (139);
		\draw [style=Edge] (144) to (141);
		\draw [style=Edge] (141) to (146);
		\draw [style=Edge] (139) to (140);
		\draw [style=Edge] (115) to (116);
		\draw [style=red edge] (151) to (153);
		\draw [style=blue edge, bend left] (151) to (153);
		\draw [dotted](174.center) to (175.center);
		\draw [style=red edge] (132) to (136);
		\draw [style=red edge] (150) to (151);
		\draw [style=blue edge, bend right=45, looseness=1.25] (150) to (151);
		\draw [dashed][style=Edge] (129) to (133);
		\draw [style=Edge] (140) to (144);
	\end{pgfonlayer}
\end{tikzpicture}
  \vspace{-1em}
  \caption{\label{fig:diagram} Influence path diagram in a NA task for the 2-layer LSTM model. The red path shows the path with the greatest attribution (the primary path) from the subject; The blue path shows the primary
    path from the intervening noun. \pxm{Can we indicate sparse/dense connections differently?}
    }
\end{figure}

\paragraph{Gate-level and Neuron-level Paths}
We define the set of gate-level nodes 
to include:
$\set{f_t^l,~ i_t^l,~ o_t^l,~ c_t^l,~ \tilde{c}_t^l,~ h_t^l~ :~ 0\leq t<T,~ 0\leq l<L}$, where T is the number of time steps (words) and L is number of LSTM layers. 
The node set also includes an attribution-specific input node ($\seq{w} - \seq{w}^0_t$) and an output node (the QoI).
An example of this is illustrated in Figure~\ref{fig:diagram}.
We exclude intermediate calculations (the solid nodes of Figure~\ref{fig:diagram}, such as $
f_{t} \odot c_{t-1}$) as their inclusion does not change the set of paths in a graph. We can also break down each vector node into scalar components and further decompose the gate-level model into a neuron-level one: 
$
\{f_{ti}^l,~ i_{ti}^l,~ o_{ti}^l,~
  c_{ti}^l,~ \tilde{c}_{ti}^l,~ h_{ti}^l~ :~ 0\leq t<T,~ 0<i<H,~  0\leq l<L\}
$, where $ H $ is the size of each gate vector. This decomposition results in an exponentially large number of paths. However, since many functions between gates in an LSTM are element-wise operations, neuron-level
connections between many neighboring gates are sparse.


\paragraph{Path Refinement}
\pxm{Describe refining a gate-level path into neuron-level paths.}
While the neuron-level path decomposition can theoretically be performed on the
whole network, in practice we choose to specify a gate-level path first, then further decompose
that path into neuron-level paths. We also collapse selected vector nodes,
allowing us to further localize a concept on a neuron level while avoiding an explosion in the number of
paths. The effect of this pipeline will be empirically justified in Section
\ref{sec:evaluation}. \pxm{Details of what this means mathematically is missing.}

\pxm{Describe taking only the sparse neuron-level paths.}

\section{Evaluation}\label{sec:evaluation}
In this section we apply influence path decomposition to the NA task.
We investigate major gate-level paths and their influence concentrations in
Section~\ref{sec:evaluation-primary}.
We further show the relations between these paths and the paths carrying grammatical number from
intervening nouns (i.e.
attractors \& helpful nouns) in Section~\ref{sec:evaluation-primary-attractors}.
In both we also investigate high-attribution neurons along primary paths allowing us to
compare our results to prior work.
\subsection{Dataset and Model}\label{sec:evaluation-dataset}
We study the exact combination of language model and NA datasets used in
the closely related prior work of \citet{lakretz2019emergence}.
The pre-trained language model of \citeauthor{Gulordavaa} and \citeauthor{lakretz2019emergence} is a 2-layer
LSTM trained from Wikipedia articles.
The number agreement datasets of \citeauthor{lakretz2019emergence} are several synthetically generated
datasets varying in syntactic structures and in the number of nouns between the subject and verb.

For example, \textit{nounPP} refers to sentences containing a
noun subject followed by a prepositional phrase such as in Figure~\ref{fig:overview}.
Each NA task has subject number (and intervening noun number if
present) realizations along singular (S) and plural (P) forms.
In listings we denote subject number (S or P) first and additional noun (if any) number second.
Details including the accuracy of the model on the NA tasks are
summarized by \citet{lakretz2019emergence}.
Our evaluation replicates part of Table 2 in said work.

\subsection{Decomposing Number Agreement}\label{sec:evaluation-primary}

We begin with the attribution of subject number on its corresponding verb, as decomposed per
Definition~\ref{def:decomposition}.
Among all NA tasks, the gate-level path carrying the most attribution is one following the
same pattern with differences only in the size of contexts.
With indices $ t $ and $ t + n $ referring to the subject and verb respectively, this path, which we
term the \emph{primary path of subject-verb number agreement}, is as follows:
\begin{equation*}
 x_t(DoI) \cdot \tilde{c}^{0} \cdot c^{0} \cdot h^0 \cdot \tilde{c}^1 \cdot \paren{c^1}^* \cdot h^1 \cdot QoI   
\end{equation*}
The primary path is represented by the red path in Figure \ref{fig:diagram}.
The influence first passes
through the temporary cell state $\tilde{c}^{0}$, the only
non-sigmoid cell states capable of storing more information than sigmoid gates, since $i, f, o \in (0,1) $ while the $tanh$ gate $\tilde{c} \in (-1,1)$.
Then the path passes through $c^{0}$, $h^0$, and similarly to ${c^1}$ through $\tilde{c}^{1}$ , jumping from the
first to the second layer.
The path then stays at ${c^1}$, through the direct connections between cell states of neighbouring
time steps, as though it is ``stored'' there without any interference from subsequent words.
As a result, this path is intuitively the most efficient and simplistic way for the model to encode and store a
``number bit.''
\pxm{Why is it intuitive and efficient and simplistic?}

The extent to which this path can be viewed as \emph{primary} is measured by two
metrics. The results across a subset of syntactic structures and number conditions mirroring those in
\citet{lakretz2019emergence} are shown in Table \ref{tab:primary}. We include 3 representative variations of the task. 
The metrics are:
\begin{enumerate}
\item $t$-value: probability that a given path has greater attribution than a uniformly sampled path on a uniformly sampled sentence. 
\item Positive/Negative Share ($\pm$Share): expected (over sentences) fraction of total positive (or negative) attribution assigned to the given positive (or negative) path.
\end{enumerate}
Per Table~\ref{tab:primary} (From Subject, Primary Path), we make our first main observation:

\begin{observation}
The same one primary path consistently carries the largest amount positive attribution across all contexts as compared to all other paths.
\end{observation}

\noindent Even in the case of its smallest share (nounPPAdv), the 3\% share is large when taking into account more than 40,000 paths in total.
Sentences with singular subjects (top part of Table~\ref{tab:primary}) have a slightly stronger concentration of
attribution in the primary path than plural subjects (bottom part of Table~\ref{tab:primary}), possibly due to English plural (infinitive) verb forms occurring more frequently than singular forms, thus less concentration of attribution is needed due to the ``default signal'' in place.

\begin{table*}[]
\centering
\resizebox{\textwidth}{!}{%
\begin{tabular}{l|l|l|l|l|l|l|l|l|l|l|l|l|l}

\multirow{3}{*}{Task} & \multirow{3}{*}{C} & \multicolumn{6}{l|}{From Subject} & \multicolumn{6}{l}{From Intervening Noun} \\\cline{3-14}
& & \multirow{2}{*}{$P_+$} & \multirow{2}{*}{$|P|$} & \multicolumn{2}{l|}{Primary Path} & \multicolumn{2}{l|}{Primary Neuron}
& \multirow{2}{*}{$P_+$} & \multirow{2}{*}{$|P|$} & \multicolumn{2}{l|}{Primary Path} &
\multicolumn{2}{l}{Primary Neuron}\\ \cline{5-8}\cline{11-14}
 & & & &+Share & $t$ & $t_{125}$ & $t_{337}$ & & & $\pm$ Share & $t$ & $t_{125}$ & $t_{337}$ \\ \toprule
\task{Simple} & \task{S} &1.0 & 16 & 0.47 & 1.0 & 0.99 & 1.0 &-  & -&- &- & -&-  \\
\task{nounPP} & \task{SS} &1.0 & 6946 & 0.1 & 1.0 & 1.0 & 1.0 &0.82 &16 &0.31(+) & 0.9 & 0.78 & 0.98    \\
\task{nounPP} & \task{SP} &1.0 & 6946 & 0.1 & 1.0 & 1.0 & 1.0 &0.23 &16 &0.24(-) & 0.23 & 0.06 & 0.15   \\
\task{nounPPAdv} & \task{SS} &1.0 & 41561 & 0.07 & 1.0 & 1.0 & 1.0 &0.92 &152 & 0.09(+) & 0.96 & 0.85 & 1.0 \\
\task{nounPPAdv} & \task{SP} &1.0 & 41561 & 0.07 & 1.0 & 1.0 & 1.0 &0.32 &152 & 0.09(-) & 0.14 & 0.13 & 0.01 \\ \midrule
\task{Simple} & \task{P} &1.0 & 16 & 0.33 & 0.93 & 0.97 & 0.99 &- &- &- & -&- & - \\
\task{nounPP} & \task{PS} &1.0 & 6946 & 0.05 & 0.91 & 0.99 & 1.0 &0.06 &16 &0.28(-) & 0.21 & 0.22 & 0.12   \\
\task{nounPP} & \task{PP} &1.0 & 6946 & 0.05 & 0.92 & 0.99 & 1.0 &0.95 &16 &0.31(+) & 0.9 & 0.97 & 0.79  \\
\task{nounPPAdv} & \task{PS} &1.0 & 41561 & 0.03 & 0.93 & 0.99 & 1.0 &0.32 &152 &0.04(-) & 0.28 & 0.41 & 0.16   \\
\task{nounPPAdv} & \task{PP} &1.0 & 41561 & 0.03 & 0.92 & 0.99 & 1.0 &0.83 &152 &0.07(+) & 0.92 & 0.99 & 0.84  \\ \bottomrule
\end{tabular}%
}
\caption{Statistics for attribution of primary paths and neurons from the subject/intervening noun: $P_+$ is the percentage of sentences with positive input attribution. Task and $C$ columns refer to sentence structures in \citet{lakretz2019emergence}.
 $|P|$ is the total number of paths; $t$ and $\pm$Share are t-values and positive/negative share, respectively.
 For calculating $t_{125}$ and $t_{337}$ of primary neurons (125 and 337), we exclude these two neurons to avoid comparing them with each other.}
\label{tab:primary}
\end{table*}

\paragraph{Primary Neurons}
We further decompose the primary path into influence passing through each neuron.
Since only connections between second layer cell states are sparse, we only decompose the segment
of the primary path from $c_t^1$ to $c_{t+n}^1$, resulting in a total of 650 (the number of hidden units) neuron-level paths. (We leave the non-sparse decompositions for future work).
The path for neuron $i$, for example, is represented as:
\begin{equation*}
  x_t(DoI) \cdot \tilde{c}^{0} \cdot c^{0} \cdot h^0 \cdot \tilde{c}^1 \cdot \paren{c_i^1}^* \cdot h^1
  \cdot QoI
\end{equation*}
To compare the attribution of an individual neuron
with all other neurons, we employ a similar aforementioned $t$-value, where each neuron-level
path is compared against other neuron-level paths.

The results of the neuron-level analysis are shown in Table~\ref{tab:primary} (From Subject, Primary Neuron).
Out of the 650 neuron-level paths in the gate-level primary path, we discover two neurons
with consistently the most attribution (neurons 125 and 337 of the second layer).
This indicates the number concept is concentrated in only two neurons.

\paragraph{Comparison with \citet{lakretz2019emergence}}
Uncoincidentally, both neurons match the units found through ablation by
\citeauthor{lakretz2019emergence}, who use the same model and dataset (neurons 988 and 776 are
neurons 125 and 337 of the second layer). This accordance to some extent verifies that the neurons found through influence paths are functionally important.
However, the $t$-values shown in Table \ref{tab:primary} show that both neuron 125 and 337 are influential regardless of the subject number, whereas \citeauthor{lakretz2019emergence} assign a subject number for each of these two neurons due to their disparate effect in lowering accuracy in ablation experiments.  One possible reason is that the ablation mechanism used in \citep{lakretz2019emergence} assumes that a ``neutral number state'' can be represented by zero-activations for all gates, while in reality the network may encode the neutral state differently for different gates.

Another major distinction of our analysis from \citet{lakretz2019emergence} regards \textit{simple}
cases with no word between subjects and verbs. Unlike \citeauthor{lakretz2019emergence}, who claim that the two identified neurons are ``long-term neurons'', we discover that these two neurons are also the only neurons important for short-term number agreement.
This localization cannot be achieved by diagnostic classifiers used by
\citeauthor{lakretz2019emergence}, indicating that the signal can be better uncovered using influence-based
paths rather than association-based methods such as ablation.



\subsection{Decomposing from Intervening Nouns}\label{sec:evaluation-primary-attractors}

Next we focus on NA tasks with intervening nouns and make the following observation:

\begin{observation}
The primary subject-verb path still accounts for the largest positive attribution in contexts with either attractors or helpful nouns. 
\end{observation}

\noindent A slightly worse NA task performance \citep{lakretz2019emergence} in cases of attractors (\task{SP}, \task{PS})
indicates that they interfere with prediction
of the correct verb.
In contrast, we also observe that helpful nouns (\task{SS}, \task{PP}) contribute
positively to the correct verb number (although they should not from a grammar perspective).
\paragraph{Primary Path from the Intervening Noun}
We adapt our number agreement concept (Definition~\ref{def:agreement}) by focusing the DoI on the intervening noun, thereby allowing us to decompose its influence on the verb number not grammatically associated with it.
In Table~\ref{tab:primary} (From Intervening Noun) we discover a similar primary path from the intervening noun: 

\begin{observation}
Attribution towards verb number from intervening nouns follows the same primary path as the subject but is of lower magnitude and reflects either positive or negative attribution in cases of helpful nouns or attractors, respectively.
\end{observation}

\noindent This disparity in magnitude is expected since the language model possibly identifies the subject as the head
noun through the prepositions such as ``behind'' in Figure \ref{fig:overview}, while still needing to track the number of the intervening noun in possible clausal
structures. Such need is comparably weaker compared to tracking numbers of subjects, possibly because in English, intervening clauses are rarer than intervening non-clauses. Similar arguments can be made for neuron-level paths.

\begin{table}[]
\centering
\resizebox{0.48\textwidth}{!}{%
\begin{tabular}{l|l|l|l|l|l|l|l|l}
\multirow{2}{*}{Task} & \multirow{2}{*}{C} & \multicolumn{7}{l}{Compression Scheme}\\ 
& & $\compl{C_{si}}$  & $\compl{C_s}$ & $\compl{C_i}$ & $C_{si}$ & $C_s$ & $C_i$ & $C$\\ \toprule
nounPP & SS &.66 &.77 &.95 &.93 &.71 &.77 &.95 \\
nounPP & SP &.64 &.36 &.94 &.64 &.75 &.40 &.74 \\
nounPP & PS &.34 &.24 &.92 &.40 &.69 &.18 &.80  \\
nounPP & PP &.39 &.66 &.91 &.76 &.68 &.58 &.97 \\ \midrule
nounPP &mean &.51 &.51 &.93 &.68 &.70 &.48 &.87 \\ \midrule

nounPPAdv & SS &.70 &.86 &.98 &.73 &.56 &.43 &1.0 \\
nounPPAdv & SP &.70 &.43 &.99 &.50 &.60 &.27 &.88  \\
nounPPAdv & PS &.38 &.22 &.98 &.76 &.79 &.56 &.96 \\
nounPPAdv & PP &.39 &.67 &.98 &.84 &.83 &.76 &1.0 \\ \midrule
nounPPAdv &mean  &.54 &.55 &.99 &.71 &.69 &.50 &.96 \\  \bottomrule

\end{tabular}%
}
\caption{Model compression accuracy under various compression schemes. $C$ is the uncompressed model.}
\label{tab:compression}
\end{table}

\subsection{Model Compression}\label{sec:modelcompression}

Though the primary paths are the highest contributors to NA tasks, it is possible that collections of associated non-primary paths account for more of the verb number concept. We gauge the extent to which the primary paths alone are responsible for the concept with compression/ablation experiments.
We show that the computations relevant to a specific path alone are sufficient in maintaining performance for the NA task. We \emph{compress} the model by specifying node sets to preserve, and intervene on the activations of all other nodes by setting their activations to constant expected values (average over all samples). We choose the expected values instead of full ablation (setting them to zero), as ablation would nullify the function of Sigmoid gates.
For example, to compress the model down to the red path in Figure~\ref{fig:diagram},
we only calculate the activation for gates $\tilde{c}_t^0$ and $\tilde{c}_t^1$ for each sample, while setting the activation of all other $\tilde{c}, f, o, i$ to their average values over all samples.
In Table \ref{tab:compression}, we list variations of the compression schemes based on the following preserved node sets:
\begin{align*}
  & C \defeq \set{f_t^l, i_t^l, o_t^l, \tilde{c}_t^l : t_{\text{sub}}<t<t_{\text{verb}}, l \in \set{0,1}}\\
  & C_s \defeq \set{\tilde{c}_{t_{\text{sub}}}^0,\tilde{c}_{t_{\text{sub}}}^1 } \;\;\; C_i \defeq \set{\tilde{c}_{t_{\text{int}}}^0, \tilde{c}_{t_{\text{int}}}^1} \\
  & C_{si} \defeq C_s \cup C_i    
\end{align*}

\noindent For example, column $C_{si}$ in Table \ref{tab:compression} shows the accuracy when the compressed model only retains the primary path from both the subject and the intervening noun while the computations of all other paths are set to their expected values; while in $\compl{C_{si}}$, all paths but the paths in $C_{si}$ are kept.

We observe that the best compressed model is $\compl{C_i}$, where the primary path from the intervening noun is left out; it performs even better than the original model; the increase comes from the cases with attractors (\task{PS}, \task{SP}). This indicates that eliminating the primary path from the attractor improves the model. The next best models apart from $C$ are $C_s$ and $C_{si}$, where primary paths are kept. Compressed models without the primary subject-verb path ($\compl{C_{si}}$,  $\compl{C_s}$, $C_i$) have performances close to random guessing.

\begin{observation}
Accuracy under path-based model compression tests corroborate that primary paths account for most of the subject number agreement concept of the LSTM.
\end{observation}

\noindent By comparing the \task{SP} and \task{PS} rows of  $\compl{C_{si}}$, $\compl{C_s}$, $C_s$, and $C_i$, we observe the effect of attractors in misguiding the model into giving wrong predictions. Similarly, we see that helpful nouns (\task{SS}, \task{PP}) help guide the models to make more accurate predictions, though this is not grammatically justified.


\section{Conclusions}\label{sec:discussion}

The combination of finely-tuned attribution and gradient decomposition lets us investigate the
handling of the grammatical number agreement concept attributed to paths across LSTM components. The concentration of attribution to a primary path and two primary cell
state neurons and its persistence in a variety of short-term and long-term contexts, even with
confounding attractors, demonstrates that the concept's handling is, to a large degree, general and
localized. Though the heuristic decisioning aspect of an LSTM is present in the
large quantities of paths with non-zero influence, their overall contribution to the concept is
insignificant as compared to the primary path. Node-based compression results further corroborate
these conclusions.


We note, however, that our results are based on datasets exercising the agreement concept in
contexts of a limited size. We speculate that the primary path's attribution diminishes with the length of the context, which would suggest that at some context size, the handling of number will devolve to be mostly
heuristic-like with no significant primary paths. Though our present datasets do not pose computational problems, the number of paths, at both the neuron and the gate level, is exponential with respect to context size. Investigating longer contexts, the diminishing dominance of the primary path, and the requisite algorithmic
scalability requirements are elements of our ongoing work.

We also note that our method can be expanded to explore number agreement in more complicated sentences with clausal structures, or other syntactic/semantic signals such as coreference or gender agreement.


\paragraph{Acknowledgement}
This work was developed with the support of NSF grant CNS-1704845 as
well as by DARPA and the Air Force Research Laboratory under
agreement number FA8750-15-2-0277. The U.S. Government is authorized
to reproduce and distribute reprints for Governmental purposes not
withstanding any copyright notation thereon. The views, opinions,
and/or findings expressed are those of the author(s) and should not
be interpreted as representing the official views or policies of
DARPA, the Air Force Research Laboratory, the National Science
Foundation, or the U.S. Government.
We gratefully acknowledge the support of NVIDIA Corporation with the donation of the Titan V GPU used for this work.



\pxm{No more than 8 pages before refs.}

\newpage

\bibliographystyle{acl_natbib} \bibliography{references}

\newpage


\end{document}